\documentclass{article}
\usepackage{graphicx} 
\usepackage{natbib}
\usepackage{amsmath}
\usepackage{parskip}
\usepackage{hyperref}

\usepackage{xcolor}
\definecolor{dark-red}{rgb}{0.4,0.15,0.15}
\definecolor{dark-blue}{rgb}{0.15,0.15,0.4}
\definecolor{medium-blue}{rgb}{0,0,0.5}
\hypersetup{
  colorlinks, linkcolor={dark-blue},
  citecolor={dark-blue}, urlcolor={medium-blue}
}

\usepackage[margin=1.2in]{geometry}

\title{Response to Promises and Pitfalls of Deep Kernel Learning}
\author{\normalsize \textbf{Andrew Gordon Wilson$^1$, Zhiting Hu$^2$, Ruslan Salakhutdinov$^{3,4}$, Eric P. Xing$^{3,5}$} \\ 
\normalsize
$^1$New York University, $^2$UC San Diego, $^3$Carnegie Mellon University, $^4$Meta, $^5$MBZUAI}

\date{}
\begin{document}

\maketitle

\begin{abstract}
This note responds to ``Promises and Pitfalls of Deep Kernel Learning'' \citep{ober2021promises}. The marginal likelihood of a Gaussian process can be compartmentalized into a data fit term and a complexity penalty. \citet{ober2021promises} shows that if a kernel can be multiplied by a signal variance coefficient, then reparametrizing and substituting in the maximized value of this parameter sets a reparametrized data fit term to a fixed value. They use this finding to argue that the complexity penalty, a log determinant of the kernel matrix, then dominates in determining the other values of kernel hyperparameters, which can lead to data overcorrelation. By contrast, 
we show that the reparametrization in fact introduces another data-fit term which influences all other kernel hyperparameters.
Thus, a balance between data fit and complexity still plays a significant role in determining kernel hyperparameters.
\end{abstract}
\vspace{2mm}

\emph{Deep Kernel Learning} (DKL) \citep{wilson2016deep} is a popular procedure, with applications ranging widely, across biological sequence design \citep{stanton2022accelerating}, 
physics informed machine learning \citep{karniadakis2021physics}, materials science \citep{choudhary2022recent}, computational chemistry \citep{keith2021combining}, autonomous vehicles \citep{al2017learning}, semi-supervised learning for predicting poverty from satellite images \citep{jean2018semi}, few-shot learning \citep{patacchiola2020bayesian}, healthcare \citep{li2021deep, chen2020deep}, probing for alien life on extraterrestrial surfaces \citep{zhu2023few}, and beyond! It has become a particularly compelling approach to Bayesian deep learning, by providing uncertainty with only a single forward pass through the network. It has played a key role in a vibrant area of research on hybrid methods \citep[e.g.,][]{calandra2016manifold, bradshaw2017}, leading to popular extensions to DKL \citep[e.g.,][]{patacchiola2020bayesian, wang2022physics}.

DKL works by providing a scalable mechanism to transform the inputs of the base kernel of a Gaussian process with a neural network. The idea marries the non-parametric flexibility and uncertainty representation of Gaussian processes with the representation learning ability of neural networks, and was introduced at a crucial time when the kernel and neural network communities were not often engaging with one another directly. The resulting deep kernel can be estimated in a variety of ways, ranging from pre-training and then freezing the network, warm-start marginal likelihood optimization, end-to-end marginal likelihood training, stochastic variational estimation, to conditional marginal likelihood optimization, amongst others. As always with a method involving many parameters and a sophisticated objective, care must go into estimation. The preferred estimation approach will depend on the architecture, application, and data, and will often achieve compelling practical performance.

In ``Promises and Pitfalls of Deep Kernel Learning'' \citep{ober2021promises} it is argued that deep kernel learning can in some cases overfit the marginal likelihood objective function, leading to poor predictive performance. However, the foundation of their argument has a key technical oversight. 

Suppose we have a dataset of input-output pairs, $\{(x_i,y_i)\}_{i=1}^N$. We can define a kernel on the inputs $k(x,x')$ that defines the covariance between the datapoints at any pair of inputs $x$ and $x'$. Evaluating the kernel on all input pairs yields the $n \times n$ covariance matrix $K_{ij} = k(x_i,x_j)$. For a Gaussian process regression model $y(x) = f(x) + \epsilon(x)$, where $f(x) \sim \mathcal{GP}(0,k)$ and $\epsilon(x) = \mathcal{N}(0,\sigma_n^2)$, the marginal likelihood of the data is
\begin{align}
\log p(\mathbf{y}) = \log \mathcal{N}(0,K) 
= -\overbrace{\frac{1}{2}\log|K+\sigma_n^2I|}^{\text{complexity penalty}} -
\overbrace{\frac{1}{2}\mathbf{y}^{\top}(K+\sigma_n^2)^{-1}\mathbf{y}}^{\text{data fit}} + \text{ c}
\end{align}
Suppose we have a kernel that can be written as a scalar amplitude times another kernel, $k(x,x') = \sigma_f^2 \hat{k}(x,x')$. For example, the popular RBF kernel can be written as $k(x,x') = \sigma_f^2 \exp\left(-\frac{1}{2\ell^2}||x-x'||^2\right)$. This kernel has a lengthscale hyperparameter $\ell$ which controls how correlated the function values are, since $k(x,x')=\text{cov}(f(x),f(x'))$. Larger lengthscales mean more slowly varying functions. \citet{ober2021promises} show that by re-parametrizing the marginal likelihood in terms of $\hat{k}$, and $\sigma_n^2 = \hat{\sigma}_n^2\sigma_f^2$, maximizing with respect to $\sigma_f^2$ to find $\hat{\sigma}_f^2 = \frac{1}{N}\mathbf{y}^{\top}(K+\hat{\sigma}_n^2I_N)^{-1}\mathbf{y}$, and substituting $\hat{\sigma}_f$ into the marginal likelihood, the reparametrized data fit term becomes $-\frac{N}{2}$ (their Proposition 1). After substituting into the marginal likelihood, they write that the remaining terms are
\begin{align}
\frac{1}{2}\log |K+\sigma_n^2 I_N| = \frac{N}{2} \log \sigma_f^2 + \frac{1}{2}\log|\hat{K}+\hat{\sigma}_n^2 I_N|
\label{eqn: ober}
\end{align}
They then argue:
\begin{quotation}
\emph{There is little freedom in minimizing $\sigma_f$, because that would compromise the data fit. Therefore, the main mechanism for minimizing the complexity penalty would be through minimizing the second term.  One way of doing this is to correlate the input points as much as possible: if there are enough degrees of freedom in the kernel, it is possible to ``hack'' the Gram matrix so that it can do this while minimizing the impact on the data fit term.} \citet[][p. 5]{ober2021promises}
\end{quotation}
However, this argument would seem to apply to virtually any popular kernel, not just a particularly flexible kernel, or deep kernel learning. For example, in the RBF kernel above, we can minimize $\log|K|$ by simply making the lengthscale $\ell$ very large, while maintaining a perfect data fit. The log determinant is the sum of log eigenvalues of the covariance matrix $K$, and as we make the lengthscale $\ell$ larger, the entries of $K$ become more similar, the matrix becomes closer to singular, and the eigenvalues decrease. But, in practice, we do not learn particularly large lengthscales, even in noise-free regression where we demand the Gaussian process fit the data as closely as possible. If we were to always learn large lengthscales in RBF kernels, then standard GP fits would look almost like straight lines!

So how can we explain this discrepancy between what is suggested by their argument, and what happens in practice? \textbf{The key missing detail in their argument is that $\sigma_f^2$ implicitly depends on the other parameters of the kernel $\theta$ and the data $\mathbf{y}$. In fact, 
through the reparametrization, a data fit term has been implicitly re-introduced!} Let's make this dependence explicit. The derived maximum value for 
\begin{align}
\hat{\sigma}_f^2(\theta) = \frac{1}{N} 
\mathbf{y}^{\top}(\hat{K}_{\theta} + \hat{\sigma}_n^2 I_N)^{-1} \mathbf{y}.
\label{eqn: sigma}
\end{align}
Now let us substitute this expression in Equation~\eqref{eqn: sigma} back into the expression of \citet{ober2021promises} of Equation~\eqref{eqn: ober}:
\begin{align}
\frac{1}{2}\log |K+\sigma_n^2 I_N| = \frac{N}{2} \log(\frac{1}{N} 
\mathbf{y}^{\top}(\hat{K}_{\theta} + \hat{\sigma}_n^2 I_N)^{-1}\mathbf{y) }+ \frac{1}{2}\log|\hat{K}_{\theta}+\hat{\sigma}_n^2 I_N|
\end{align}
We can see now that it is not clear that the original $-\log|K_{\theta}+\sigma_n^2I|$ would be made arbitrarily large, since other terms in the marginal likelihood still have a complex dependence on $\theta$. In fact, \textbf{there is a data fit term outside of the $\log|K|$.} It is also not clear, from this formulation, that maximizing this expression with a flexible kernel is going to strongly overcorrelate the data: indeed an overly large length-scale in the RBF kernel, which would minimize $\log|K|$, is prevented by the other terms.

\textbf{The key oversight in \citet{ober2021promises} is the implication that $\hat{\sigma}_f^2$ is fixed.} It is not. It depends on the parameters of the kernel, and the data. It is not clear that the ``main mechanism for minimizing the complexity penalty would be through minimizing the second $\log|\hat{K}+\hat{\sigma}_n^2|$ term''.

More generally, while it is well known that the marginal likelihood does automatically calibrate the complexity of the model \citep[e.g.,][Chapter 28]{mackay2003information}, it can of course be misaligned with generalization, as with any likelihood. We argue in \citet{lotfi2022bayesian} the various different ways in which the marginal likelihood specifically can be misaligned with generalization. One way is overfitting due to lack of uncertainty representation. Another, not discussed in \citet{ober2021promises}, is underfitting, where certain parameter settings $\theta$ lead to a distribution over functions unlikely to generate the training data, but where posterior contraction can lead to good generalization. This type of underfitting can be mitigated by instead maximizing a conditional log marginal likelihood (CLML). \citet{lotfi2022bayesian} shows in Figure 9 that the CLML can improve the performance of DKL over LML optimization, especially on problems with smaller numbers of data points. 

At the same time, it is being seen that a compression bias, implicitly induced through scale, can play a major role in the generalization performance of large neural networks \citep{wilson2025deep}. It could therefore be desirable to make this compression bias more explicit in the objective functions we use to train our models. The marginal likelihood provides one compelling mechanism for encoding such a bias, since it encourages minimum description length solutions \citep{mackay2003information}. Indeed in practice it has proven itself time and again as a useful objective, in DKL and beyond.

Any time one is maximizing a sophisticated objective function with many degrees of freedom, performance will indeed be sensitive to the details. There are many configurations of DKL: end-to-end training through the marginal likelihood, warm-starting by pre-training the neural network and then fine-tuning with marginal likelihood, or simply pre-training the neural network and freezing it as input to the kernel. The best performing approach will depend on the architecture, application, and many other variables. As always, numerical stability and initialization will also be important considerations. While SVDKL \citep{wilson2016stochastic} is another successful variant of DKL, and mini-batch optimization can be computationally valuable, it is not often necessary for achieving practical success with DKL. Good performance is regularly achieved in full batch settings, for both regression and classification, even in online learning with small datasets \citep[e.g.,][]{li2024study}. Other interventions can also be useful. In addition to using the CLML objective, a fully Bayesian treatment could be helpful (though expensive, particularly with the exact marginal likelihood objective which does not factorize across datapoints). Other regularization such as weight decay could be reasonable as well.

Like with any method that involves big models and moving parts, we must be thoughtful about how we do estimation. But such procedures can be highly practical. Indeed, variational autonencoders use the marginal likelihood to train the whole decoder network, which usually involves millions of parameters \citep{kingma2013auto}! 

DKL has now significantly influenced the development of new methodological research, as well as a truly incredible spectrum of successful applications! The adoption and practical successes of DKL are still growing.
When we first worked on DKL, it was unclear what sort of impact it would have, or even what the right venue might be, since the kernel and deep learning communities were very much in tension. At the time, we wrote \emph{``we hope that this
work will help bring together research on neural networks and kernel methods, to inspire
many new models and unifying perspectives which combine the complementary advantages
of these approaches.''} We are happy that this hope has been realized much more than we anticipated. And now, with hindsight, we would like to encourage readers to similarly pursue ideas at the intersection of areas that are viewed as competing, but perhaps shouldn't be: Bayesian credible sets with conformal calibration, neurosymbolic approaches, or large models with scientifically interpretable inductive biases.

\bibliography{refs}
\bibliographystyle{apalike}
\end{document}